%% file: c2d_wacv.tex
\pdfoutput=1
\documentclass[10pt,twocolumn,letterpaper]{article}

\usepackage{wacv}
\usepackage{times}
\usepackage{epsfig}
\usepackage{graphicx}
\usepackage{amsmath}
\usepackage{amssymb}
\usepackage[accsupp]{axessibility}  % Improves PDF readability for those with disabilities.
\usepackage{cellspace}
\usepackage{amsfonts}
\usepackage{physics} 
\usepackage[protrusion=true,tracking=true,kerning=true,expansion,final]{microtype}   % microtypography
\usepackage{booktabs} % for professional tables
\usepackage{subcaption}
\usepackage{nicefrac} % nice flat fraction
\usepackage{cancel}

\usepackage{xspace}
\usepackage{upref}
\usepackage{textcomp}
\usepackage[T1]{fontenc}

\usepackage{array}
\usepackage{amssymb}
\usepackage{gensymb}
\usepackage{bbding}
\usepackage{mathtools}
\usepackage{centernot} % not parallel, etc.
\usepackage{bigints}
\usepackage{dsfont} %mathbb 1
\usepackage{esint} % beatiful integrals
\usepackage{multirow}
\usepackage{enumitem}
\usepackage{graphbox} %loads graphicx package
\usepackage{xcolor}

\usepackage{algorithm}
\usepackage{algpseudocode}
\usepackage{algorithmicx}
\usepackage[square,sort,comma,numbers]{natbib}

\usepackage{varioref} % references to show page
\usepackage[pagebackref=true,breaklinks=true,colorlinks,bookmarks=true]{hyperref}

\usepackage[capitalise]{cleveref} 
\crefname{appsec}{appendix}{appendices}
\Crefname{appsec}{Appendix}{Appendices}

\usepackage{url}

\renewcommand{\cite}[1]{\citep{#1}}
\definecolor{mydarkblue}{rgb}{0,0.08,0.45}
\definecolor{urlcolor}{rgb}{0,.145,.698}
\definecolor{linkcolor}{rgb}{.71,0.21,0.01}
\hypersetup{ %
	pdftitle={Contrast to Divide: self-supervised pre-training for learning with noisy labels},
	pdfauthor={Evgenii Zheltonozhskii, Chaim Baskin, Avi Mendelson, Alex M. Bronstein, Or Litany},
	pdfsubject={}, 
	pdfkeywords={},
	pdfborder=0 0 0,   
    pdfpagemode=UseOutlines,
	breaklinks=true, % so long urls are correctly broken across lines
	colorlinks=true,
	bookmarksnumbered=true,
	urlcolor=urlcolor,
	linkcolor=linkcolor,
	citecolor=mydarkblue,
	filecolor=mydarkblue,
	pdfview=FitH}

\renewcommand*{\backref}[1]{} % for backref < 1.33 necessary
\renewcommand*{\backrefalt}[4]{%
	\ifcase #1 %
	\or
	(cited on p. #2)%
	\else
	(cited on pp. #2)%
	\fi
}
\makeatletter
\renewcommand{\@biblabel}[1]{#1.}
\makeatother

\vbadness=\maxdimen
\usepackage{bm}
\newcommand{\eb}[1]{{\scriptsize\,$\pm$\,#1}}

\def\wacvPaperID{580} % Enter the WACV Paper ID here

\wacvfinalcopy % *** Uncomment this line for the final submission

\ifwacvfinal
\def\assignedStartPage{9876} % *** Enter the assigned starting page number (instead of 9876)
\fi

\ifwacvfinal
\else
\fi

\begin{document}

%%%%%%%%% TITLE

\title{Contrast to Divide: Self-Supervised Pre-Training for Learning with Noisy Labels}

\newcommand*\samethanks[1][\value{footnote}]{\footnotemark[#1]}
\author{Evgenii Zheltonozhskii\thanks{Equal contribution.}\hspace{4pt}$^{1}$,
Chaim Baskin\samethanks[1]\hspace{4pt}$^{1}$,
Avi Mendelson$^{1}$,
Alex M. Bronstein$^{1}$,
Or Litany$^{2}$\\
    {\tt\small $^{1}$Technion -- Israel Institute of Technology, $^{2}$NVIDIA}\\
{\tt\small \href{mailto:evgeniizh@campus.technion.ac.il}{evgeniizh@campus.technion.ac.il};
\href{mailto:chaimbaskin@campus.technion.ac.il}{chaimbaskin@campus.technion.ac.il};}\\
{\tt\small \href{mailto:mendlson@technion.ac.il}{mendlson@technion.ac.il};
\href{mailto:bron@cs.technion.ac.il}{bron@cs.technion.ac.il};}\\
{\tt\small \href{mailto:orlitany@gmail.com}{orlitany@gmail.com}}

}

\maketitle
%\thispagestyle{empty}

%%%%%%%%% ABSTRACT
\begin{abstract}
The success of learning with noisy labels (LNL) methods  relies heavily on the success of a warm-up stage where standard supervised training is performed using the full (noisy) training set. 
In this paper, we identify a ``warm-up obstacle'': the inability of standard warm-up stages to train high quality feature extractors and avert memorization of noisy labels. 
We propose ``Contrast to Divide'' (C2D), a simple framework that solves this problem by pre-training the feature extractor in a self-supervised fashion. 
Using self-supervised pre-training boosts the performance of existing LNL approaches by drastically reducing the warm-up stage's susceptibility to noise level, shortening its duration, and improving extracted feature quality. 
C2D works out of the box with existing methods and demonstrates markedly improved performance, especially in the high noise regime, where we get a boost of more than 27\% for CIFAR-100 with 90\% noise over the previous state of the art.
In real-life noise settings, C2D trained on mini-WebVision outperforms previous works both in WebVision and ImageNet validation sets by 3\% top-1 accuracy. 
We perform an in-depth analysis of the framework, including investigating the performance of different pre-training approaches and estimating the effective upper bound of the LNL performance with semi-supervised learning.
Code for reproducing our experiments is available at \url{https://github.com/ContrastToDivide/C2D}. 
\end{abstract}

%%%%%%%%% BODY TEXT
\input{sections/010_intro}

\input{sections/020_related}
\input{sections/030_method}
\input{sections/040_experiments}
\input{sections/050_details}
\input{sections/060_conclusion}

{\small
\bibliographystyle{ieee_fullname}
\bibliography{c2d_wacv}
}

%\clearpage
%%%%%%%%% Appendix
\input{sections/070_appendix}

\end{document}

%% file: sections/010_intro.tex
\section{Introduction}
\label{sec:intro}
\begin{figure*}
    \centering
    \begin{tabular}{l  ll r} 
    &\multicolumn{1}{c}{$20\%$ noise}&\multicolumn{1}{c}{$90\%$ noise} & \multicolumn{1}{c}{Self-supervised}\\
    \rotatebox[origin=c]{90}{C2D}
    &\includegraphics[align=c,width=0.27\textwidth]{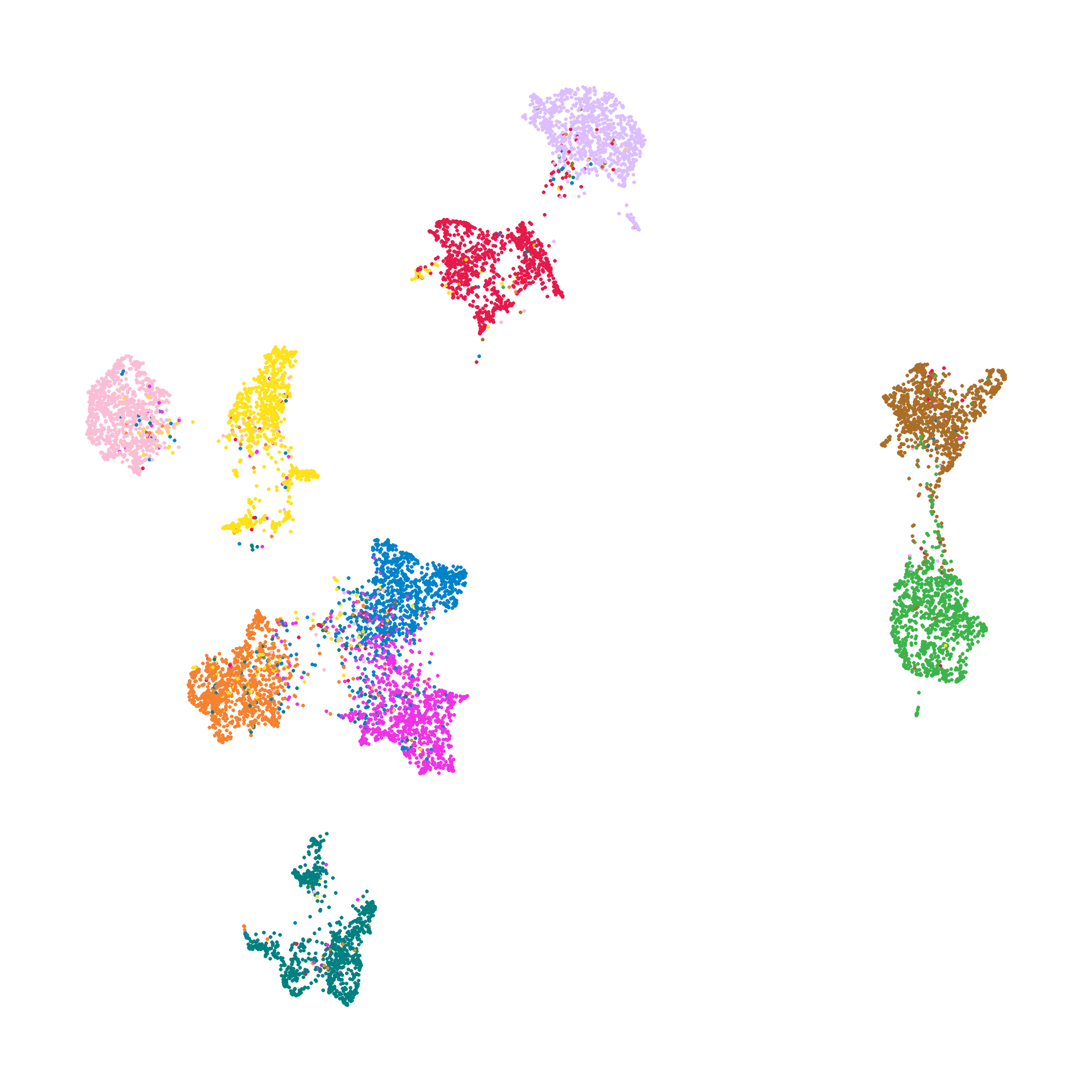}&
    \includegraphics[align=c,width=0.27\textwidth]{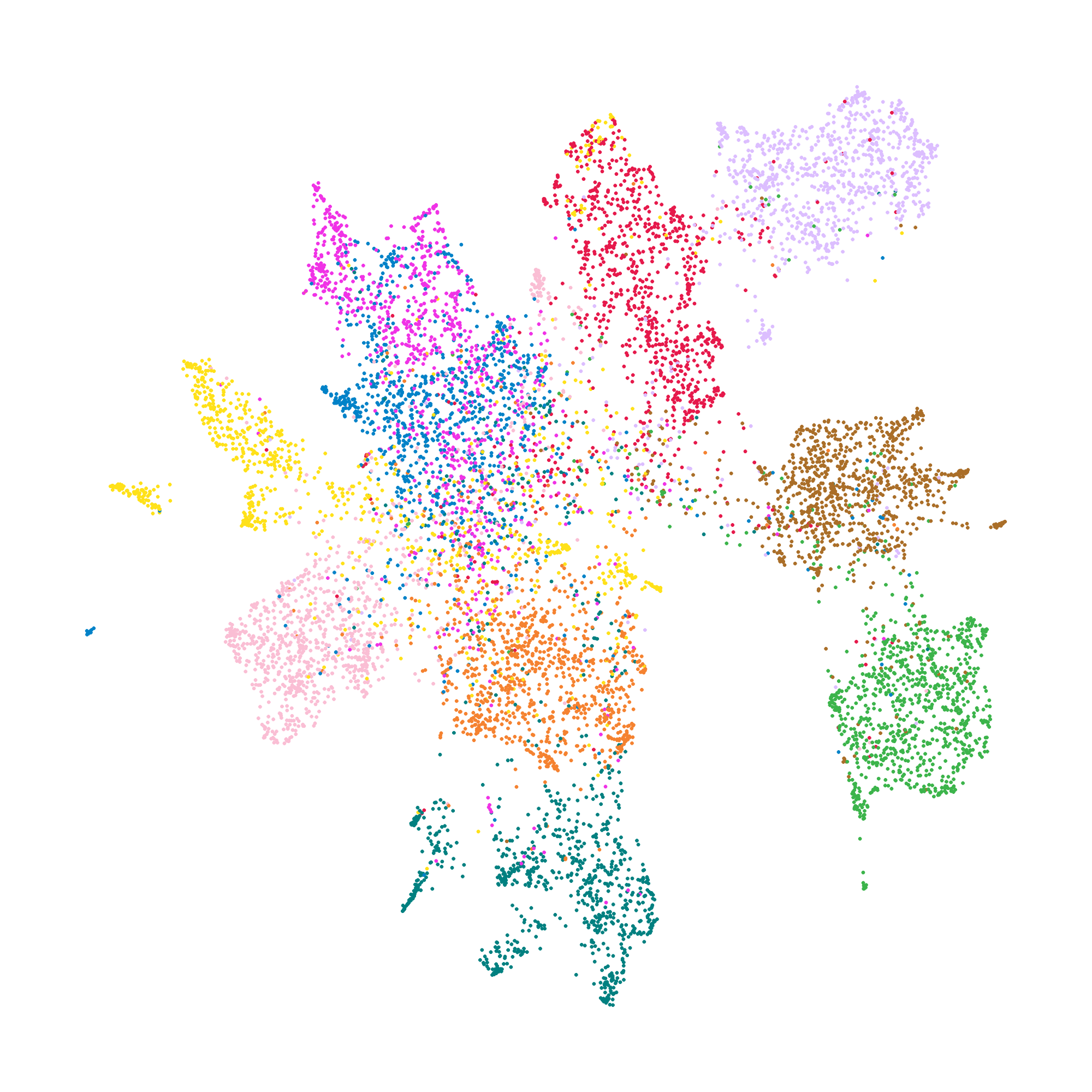}&
    \includegraphics[align=c,width=0.27\textwidth]{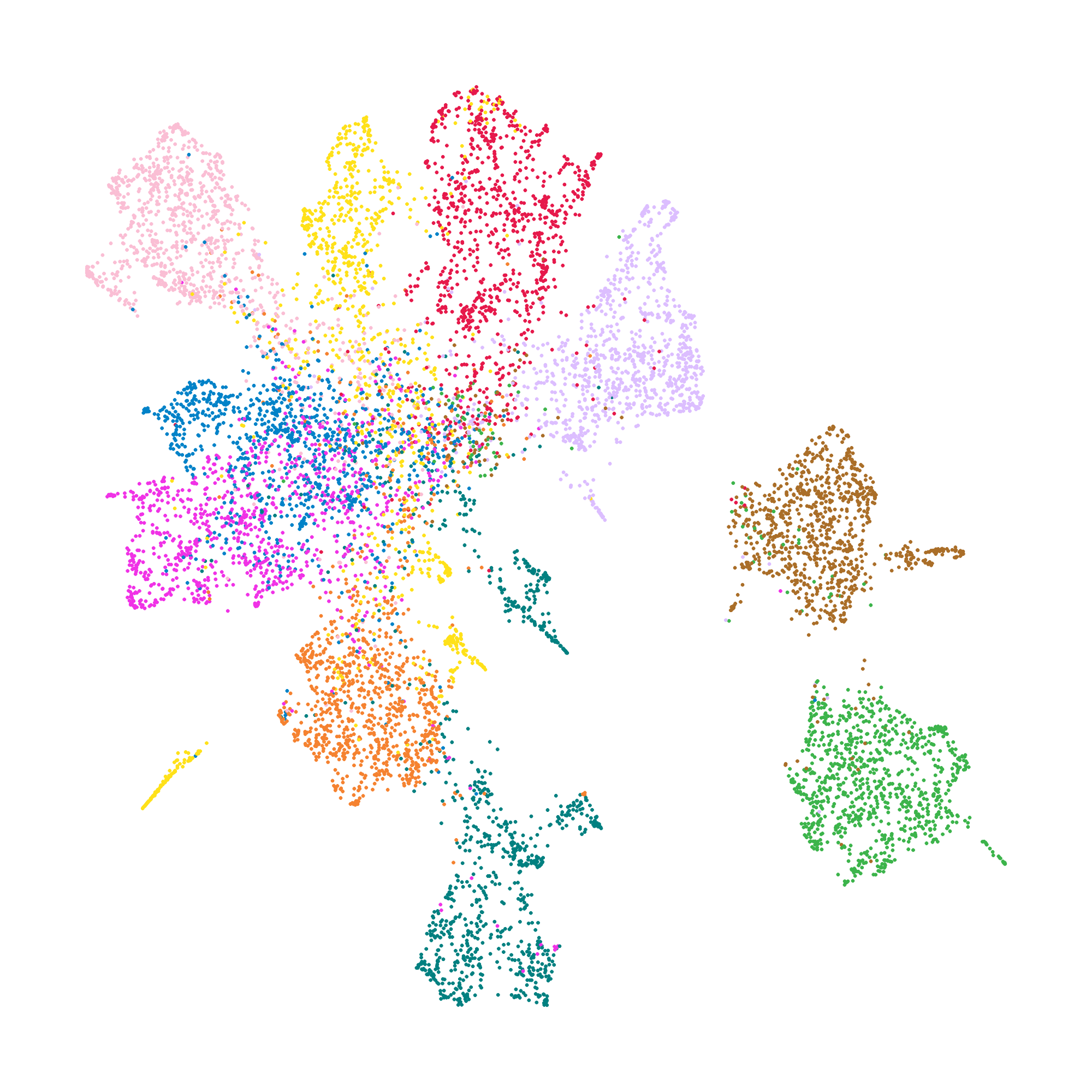}
    
    \\
    \rotatebox[origin=c]{90}{DivideMix}&\includegraphics[align=c,width=0.27\textwidth]{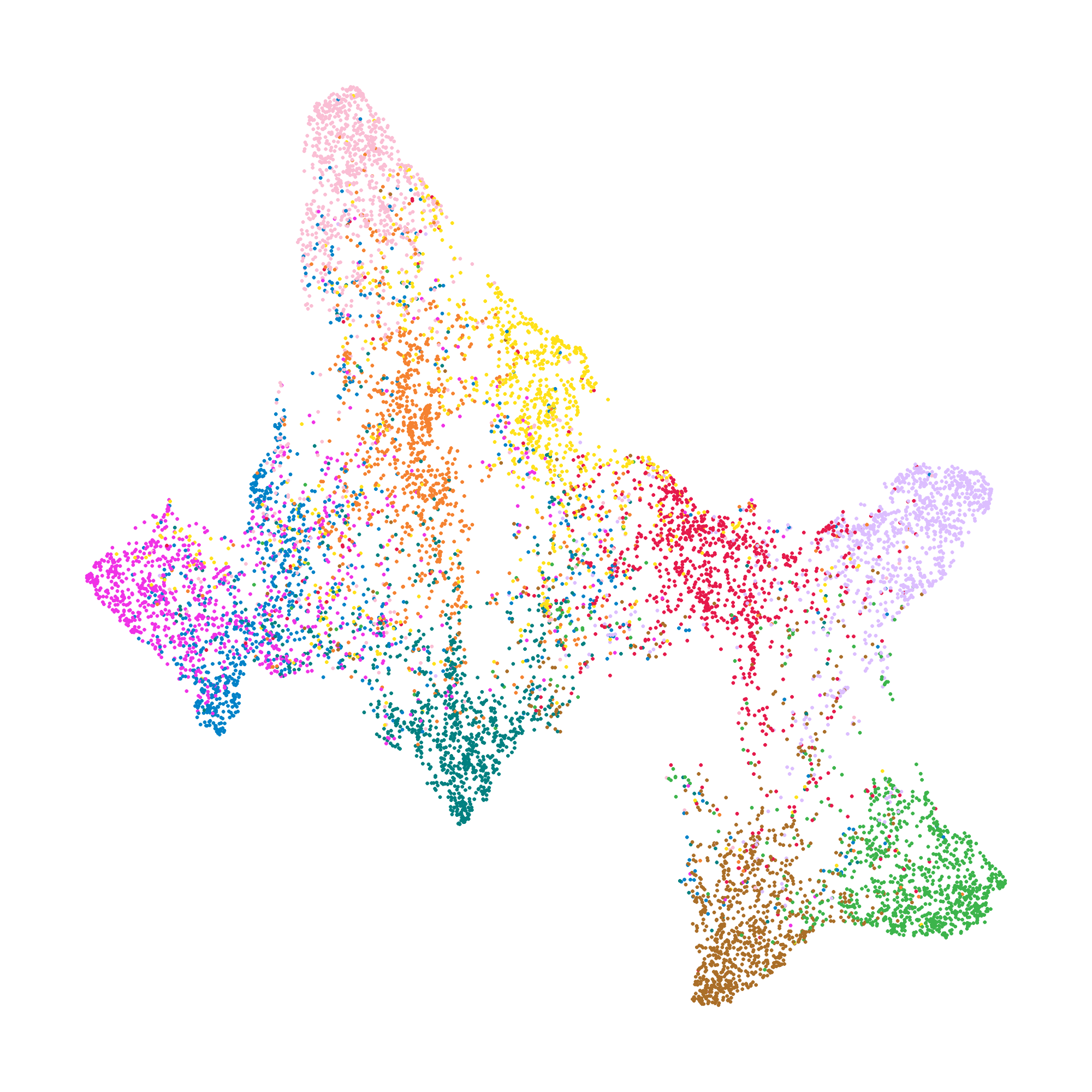}&
    \includegraphics[align=c,width=0.27\textwidth]{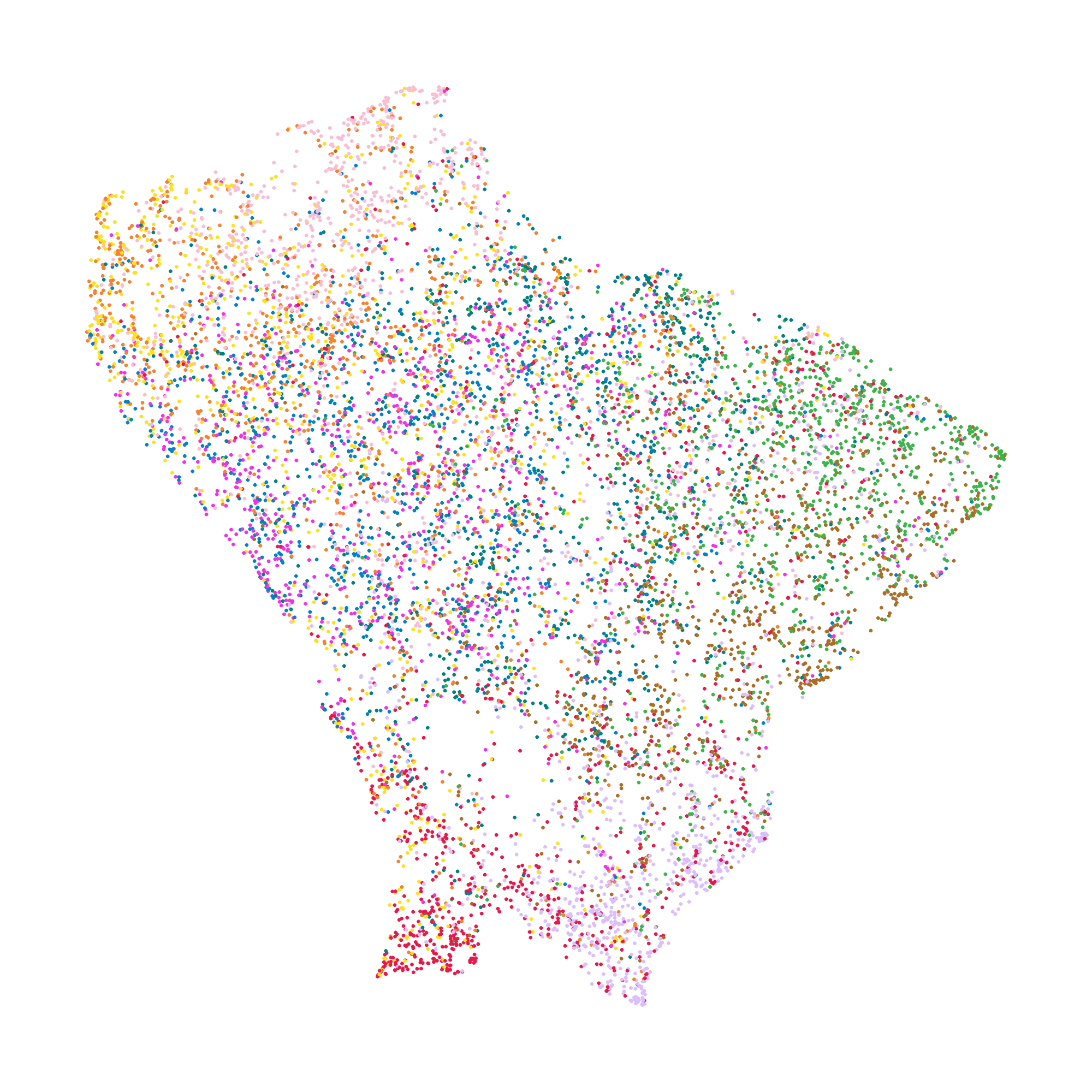}&
    \includegraphics[align=c,width=0.15\textwidth]{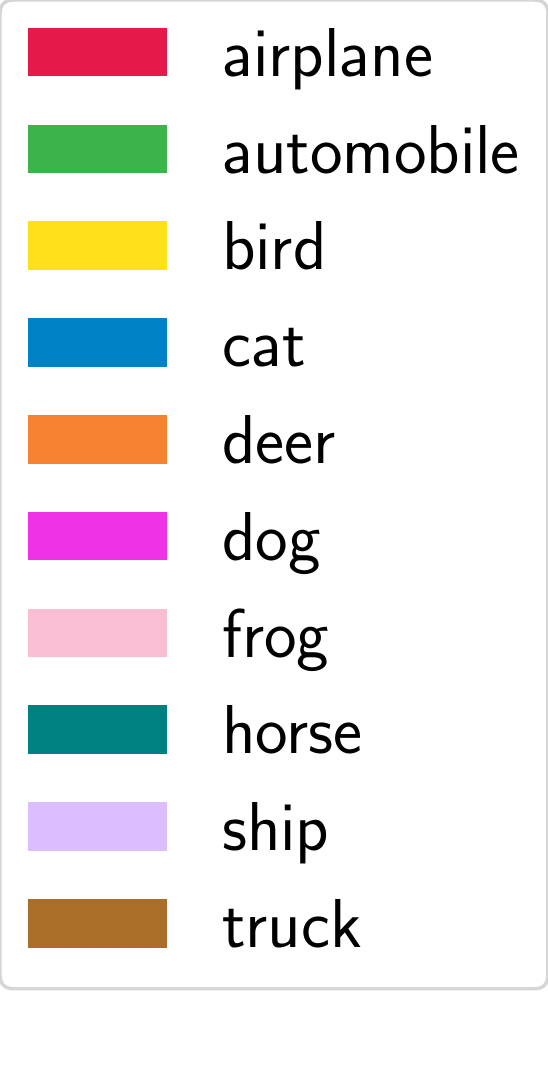}
    \end{tabular}
    \caption{A UMAP \cite{lel2018umap} of features extracted from CIFAR-10 using C2D (top row) vs. DivideMix (bottom row) for 20\% and 90\% noise at the end of warm-up stage, as well as self-supervised pre-training. Colors indicate the ground-truth labels.}
    \label{fig:umap}
\end{figure*}

Many deep-learning-based methods owe their success to the availability of large data sources with reliable labels.  Quality annotation at scale, however, is often prohibitively expensive. 
Two common approaches that address this challenge are semi-supervised learning and learning with noisy labels (LNL). The former assumes the availability of a limited amount of high-quality labeled data as well as a large amount of unlabeled data of the same distribution. The main challenge is to propagate the labels to the unlabeled samples to allow gleaning knowledge from them as well.  In contrast, the latter approach suggests acquiring cheap annotations at scale at the cost of having some mislabeled data. Examples of such processes include web crawling \cite{xiao2015learning,li2017webvision}, automatic annotation based on meta-data \cite{mahajan2018limits}, and uncurated crowdsourcing \cite{kuznetsova2020open}.  
Though seemingly different, the two approaches are in fact closely related. Many semi-supervised learning approaches are based on predicting pseudo-labels for the unlabeled data, which are, effectively, noisy labels. From the other end, an LNL setting can be converted  into a semi-supervised one by identifying and discarding the noisy labels. Separation of the noisy labeled samples from the clean ones is one of the key challenges in LNL. 

To that end, multiple LNL methods utilize a ``warm-up'' stage \cite{patrini2017making,han2019deepself,li2020dividemix,liu2020earlylearning} -- short supervised training on the full noisy dataset that precedes the more sophisticated algorithms designed to deal with label noise.
During warm-up, the network's inherent robustness to noise tends to lower the classification loss of the cleanly labeled samples faster than that of the noisy ones \cite{cicek2018speedas,liu2020earlylearning}. 
While this stage has not received much attention in previous works (possibly due to its algorithmic simplicity), it is in fact crucial to the success of LNL. Unfortunately, it is prone to memorizing noise, and thus its efficacy is contingent on the noise level as well as the amount of training iterations and other hyperparameters.

These limitations create a significant obstacle to improving the performance of LNL approaches.
While supervised pre-training on a large clean dataset (e.g., ImageNet \cite{ILSVRC15}) may seem to be a possible solution to this problem, the availability of such data may be limited in some domains (e.g., medical data). In addition, our experiments show that in some scenarios, ImageNet pre-training may degrade the performance of LNL algorithms.

We propose to overcome the warm-up obstacle by using \textit{unsupervised} pre-training. Building on the recent success of self-supervised learning \cite{henaff2019cpc2,chen2020simclr,tian2020infomin,zbontar2021barlow}, especially in closely related semi-supervised tasks \cite{chen2020simclr2}, we generate high-quality features by pre-training on the unlabeled train set samples.  Thus, we benefit simultaneously from several effects. We do not require external data sources; by ignoring the labels, we eliminate the influence of noise on the pre-training stage and prevent noise memorization;  finally, by operating on the training set, we avoid a domain gap.  Importantly, this can be seamlessly combined with any LNL method. 

Altogether, our framework provides a significant boost over previous LNL methods, with much better consistency across different noise levels. 
For example, with 90\% symmetric noise, we achieve a more than 27\% accuracy boost for CIFAR-100 with PreAct ResNet-18. 
In real-life settings, on mini-WebVision the proposed framework achieves an accuracy boost of more than 3\% on top-1 accuracy both in WebVision and ImageNet validation sets; on Clothing1M it matches performance of the ImageNet pre-training without any external data.

Below, we outline our main contributions. 
\begin{itemize}
    \item First, we identify and characterize the warm-up importance for LNL. Our proposed framework, ``Contrast to Divide'' (C2D), improves warm-up performance by utilizing self-supervised pre-training.
    \item C2D significantly outperforms state-of-the-art methods on standard benchmarks that do not utilize external data: CIFAR and WebVision. 
    Moreover, on the challenging Clothing1M benchmark, C2D matches the state of the art that uses pre-training on ImageNet.
    \item We perform an extensive analysis of C2D, including loss separation and feature quality for different initialization schemes, loss distribution after the warm-up stage, and the performance gap between C2D and semi-supervised learning.
\end{itemize}

%% file: sections/020_related.tex
\section{Related work}
\label{sec:related}

\paragraph{Self-supervised learning.}
Self-supervised learning aims to learn representations that are meaningful in some general sense, without using externally provided labels. Usually, this is done by solving a pretext task. One family of methods is based on reconstructing a corrupted version of the input  \cite{vincent2008extracting, zhang2016colorful,pathak2016context, zhang2017split}.
Other methods opt for using a classification task based on context prediction \cite{doersch2015unsupervised,noroozi2016unsupervised, gidaris2018unsupervised,kolesnikov2019revisiting} or clustering \cite{caron2018deep}. 
Nevertheless, all these methods impose an inherent problem when facing a particular downstream task that may not be well correlated with the self-supervised objective. Thus, there is no guarantee that the key information is retained and can be extracted from the features \cite{misra2020self}. Some methods  propose to remedy this problem by making the self-supervised task aware of the downstream one  \cite{zhai2019s4l,khosla2020supervised}. 

Recently, a revival in self-supervised techniques based on contrastive loss \cite{hadsell2006dimensionality} has shown markedly improved performance in large-scale computer vision tasks
\cite{henaff2019cpc2,chen2020simclr,tian2020infomin,xie2020pointcontrast}. Subsequently, similar approaches without utilizing contrast between samples were proposed \cite{grill2020bootstrap,chen2020simsiam,zbontar2021barlow}. % Most relevant to our method is the result reported by \citet{chen2020simclr2}: self-supervised features obtained by contrastive learning can improve semi-supervised classification tasks after fine-tuning.

\paragraph{Semi-supervised learning.}
Given a partially labeled dataset, semi-supervised techniques aim at utilizing the unlabeled samples for boosting the learning procedure beyond what is achievable with just the labeled set. A simple yet efficient baseline for this problem 
%which stood the test of time 
is pseudo-labeling \cite{lee2013pseudo,arazo2019pseudolabeling,xie2019self,yalniz2019billion}. In its basic form, this solution uses a network trained on the labeled subset to predict labels for the unlabeled set. These new labels, in turn, are used to refine the network (or a larger one) on the now fully labeled set. Another popular approach to semi-supervised learning is consistency regularization, where in addition to the cross-entropy loss, consistency is enforced between different perturbations of unlabeled (and possibly labeled) samples. Various implementations of those perturbation were studied, including predictions by different networks \cite{tarvainen2017meanteacher}, adversarial examples \cite{miyato2017virtual}, and augmentations \cite{xie2019unsupervised,berthelot2019mixmatch,sohn2020fixmatch,french2020milking}. Recent methods have shown competitive results on CIFAR-100 using labels for as little as 1\% of samples. Moreover, self-supervised pre-training can further improve semi-supervised classification \cite{chen2020simclr2}.

\paragraph{Learning with noisy labels.}
There are many variants of the LNL problem. While some methods \cite{veit2017learning,litany2018soseleto,zhang2020severe} assume the availability of a small subset of clean labels, we do not make those assumptions. We also consider closed-set noisy labels, i.e., where the mislabeled images belong to one of the training classes as opposed to the open-set setup \cite{wang2018openset,zhang2018generalized}.

Existing methods for LNL can be divided into two broad categories: loss modification and noise detection.  The former group includes techniques that account for noise distribution \cite{patrini2017making,xiaobo2019anchor,yao2020dual}. Alternatively, the loss itself may be replaced by a more robust version, such as mean absolute error \cite{ghost2017mae}, generalized cross-entropy \cite{zhang2018generalized}, determinant-based mutual information \cite{xu2019ldmi}, or a meta-learning objective \cite{li2019learning}. On the other hand, noise detection methods aim to discover which samples are mislabeled to either relabel or discard them. Techniques for detecting noisy labels include utilizing multiple networks in a teacher-student \cite{jiang2018mentornet} or mutual teaching \cite{han2018coteaching,yu19coteachingplus} framework, geometry \cite{han2019deepself}, mixture models \cite{arazo2019mcorr,li2020dividemix}, and quantiles of counterfactual loss distribution of samples \cite{song2020robust}. These are often based on the observation that samples with noisy labels converge slower than those with clean ones \cite{arpit2017closer,cicek2018speedas,li2019gradient,pleiss2020identifying}. Hybrid methods that try to mix both noise detection and loss modification were also proposed \cite{song2019selfie,liu2020earlylearning}.

%% file: sections/030_method.tex
\input{sections/result_table_cifar_10.tex}

\section{The warm-up obstacle}
\label{sec:warmup}
C2D is motivated by the observation of an inherent obstacle that is at the core of LNL methods. It has been shown that deep networks can perform meaningful learning in the presence of noise before they enter a memorization phase \cite{pleiss2020identifying}. LNL methods utilize this behavior by performing a warm-up -- supervised training on the full set of (noisy) labels for a short period of time. Most methods utilize either ``hard'' (starting an LNL procedure after a number of epochs \cite{li2020dividemix}) or ``soft'' (gradually increasing the weight of additional regularization terms \cite{liu2020earlylearning}) version of warm-up.

\begin{figure}
\centering
        \includegraphics[width=\linewidth]{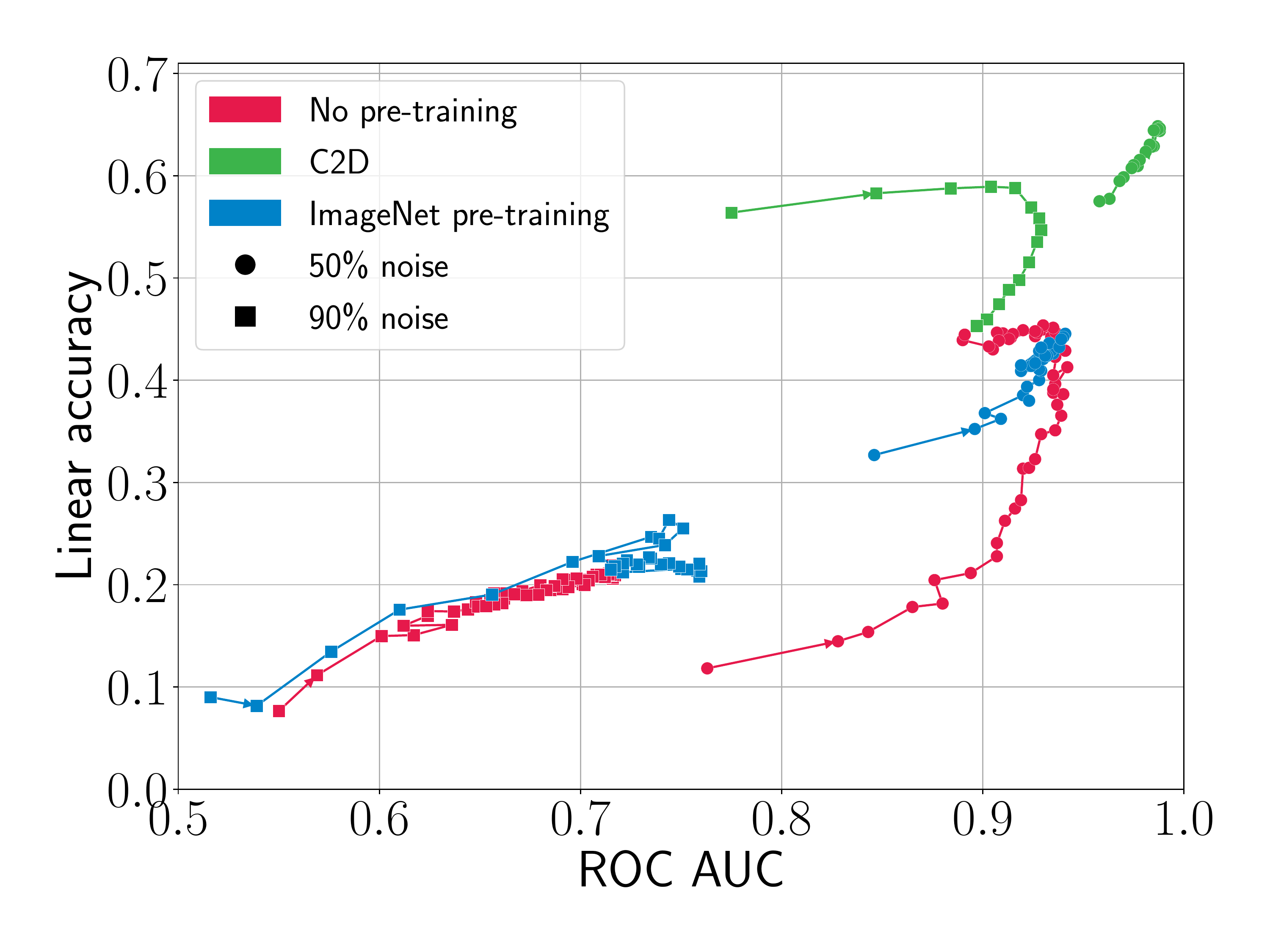}
 \caption{The ROC-AUC score of noise detection and the linear accuracy using clean labels  under various noise levels for CIFAR-100 for standard warm-up, ImageNet pre-training, and C2D. Each point is one epoch of training, arrowhead denotes time direction.  Colors denote the pre-training scheme and markers denote the noise level.}
\label{fig:tradeoff}
\end{figure}

A warm-up stage has two main  goals: loss separability and feature extraction. The former means that the model is still in the early learning phase, allowing the follow up stage to rely on noisy labeled samples having high loss values clean labeled ones having low loss values. The latter refers to the quality of representation learned by the model. 
Little, however, is understood about the \textit{determinants} of network robustness to noise or how to boost it. As a result, the warm-up performance in LNL methods is consequential and bounded by an unavoidable memorization. Current practice in LNL is to merely adjust the warm-up length according to the observed robustness of the model under different noise levels. We identify this as a major obstacle in the ability to improve performance.

To demonstrate this phenomenon, we run a supervised training on CIFAR-100 with noise and measured the level of the aforementioned properties at each epoch. The results are visualized in \cref{fig:tradeoff}. Separation is measured as the ROC AUC of the noise detection with Gaussian mixture model (GMM) applied to loss values \cite{li2020dividemix}, and feature quality is measured using the classification accuracy of a linear classifier trained without noise -- a standard approach for feature quality assessment \cite{zhang2016colorful,henaff2019cpc2,chen2020simclr}. 
As can be seen, the two measures are strongly correlated, peaking jointly at some iteration, which, of course, is unknown unless clean labels are available. 
Critically, not only do the separation and feature quality values deteriorate quickly as the noise level increases, but also no known remedy exists. In other words, even if we knew the \textit{optimal} warm-up length, the current LNL toolbox lacks the tools to improve the observed values. The effect of feature deterioration as the noise level increases can also be seen in the bottom part of \cref{fig:umap}, where we visualized the extracted features after the warm-up using UMAP \cite{lel2018umap}. %Clearly, label noise has a detrimental effect on feature separation. 

To circumvent the deterioration issue, prior works have resorted to supervised pre-training on an external dataset, such as ImageNet. This solution suffers from two disadvantages. First, it necessitates a large, cleanly-labeled dataset of a similar domain, which is not always available. Second, as will be made apparent by our analyses, features generated via supervised pre-training may fall short in noisy label separation.  
Our solution to overcoming the warm-up obstacle is to use self-supervised pre-training.

\section{Contrast to Divide}
\label{sec:method}
As discussed in \cref{sec:warmup}, the warm-up phase performance in LNL pipelines is bounded by memorization. Encouraged by the recent success in semi-supervised learning \cite{chen2020simclr2}, we study whether self-supervised pre-training could break this barrier. More specifically, given a dataset with contaminated labels, we propose a straight-forward two-phase framework. First, we perform self-supervised contrastive learning \cite{chen2020simclr,zbontar2021barlow} to obtain high-quality feature extractor (contrast phase). We then proceed with a standard LNL algorithm that can now better detect noisy labels (divide phase). Much like standard transfer learning, this framework can be used to boost virtually any existing LNL method. %The main difference lies in that
We do not, however, rely on a cleanly labeled external data source; pre-training is done directly on the target training set. Importantly, by discarding the labels, we avoid label noise influence on feature extractor and provide a robust initialization as can be seen in \cref{fig:tradeoff}. Even under extreme noise level conditions, this initialization boosts the warm-up far beyond the memorization bound. 
%We tested C2D with two state-of-the-art LNL methods, ELR+ and DivideMix, and found that in both cases the improvement was major. 
As shown in the experimental section, the improved loss separation supports both explicit separation using a classification model as well as an implicit one based on regularization terms.
%the prediction regularization of ELR+, as well as the explicit GMM-based clean/noisy separation done by DivideMix. 

%% file: sections/result_table_cifar_10.tex
\begin{table*}
	\centering
    \setlength\extrarowheight{2pt}
    \resizebox{\linewidth}{!}{
\begin{tabular}{@{\extracolsep{4pt}}l l l cccccc@{}} 
	\toprule	 
	\multirow{2}{*}{\textbf{Method}}&\multirow{2}{*}{\textbf{Architecture}}   &&\multicolumn{6}{c}{\textbf{Noise rate}}\\ \cline{4-9}
             &       &       & \textbf{20\%}  & \textbf{50\%}  & \textbf{80\%}  & \textbf{90\%}  & \textbf{95\%}  &\textbf{Asym. 40\%} \\
	\midrule
Meta-learning               & \multirow{2}{*}{PreAct ResNet-32 }      & Peak  & 92.9  & 89.3  &77.4   & 58.7  &   --  & 89.2 \\
\cite{li2019learning}       &                               & Final  & 92.0  & 88.8  & 76.1  & 58.3  &   --  & 88.6 \\
	\midrule				
ELR+                        & \multirow{2}{*}{ResNet-34 }   & Peak   &   --    &   --    &    --   &	--    &  --   & -- \\
\cite{liu2020earlylearning} &                               & Final & 95.8  & 94.8  & 93.3  & 78.7	&   --  & 93.0 \\
	\midrule		
DivideMix                   & \multirow{2}{*}{PreAct ResNet-18 }      & Peak  & 96.1  &94.6	&93.2   &76.0   &  --   &93.4\\	            
\cite{li2020dividemix}      &                               & Final  & 95.7  &94.4   & 92.9  &  75.4 &   --  & 92.1\\	 	
	\midrule		
\multirow{2}{*}{CE+mixup with SimCLR}  & \multirow{2}{*}{ResNet-34 }      & Peak  & 94.98\eb{0.17} & 93.83\eb{0.12}	 & 91.33\eb{0.17} & 87.54\eb{0.45} & 73.97\eb{1.22}   & 91.63\eb{0.38} \\	            
			                &                               & Final  & 94.56\eb{0.22} & 93.20\eb{0.25} & 90.77\eb{0.57} &   85.77\eb{2.42}  &   72.71\eb{1.39} & 90.39\eb{0.35}  \\	
	\midrule	
\multirow{2}{*}{C2D (ELR+ with SimCLR)}  & \multirow{2}{*}{ResNet-34 }      & Peak  & \textbf{96.83\eb{0.10} }&	\textbf{95.96\eb{0.09}} & 93.67\eb{0.16}& 89.94\eb{0.20} & 83.23\eb{0.68}   & \textbf{94.32\eb{0.48}}\\	            
			                &                               & Final  & \textbf{96.74\eb{0.12}}  & \textbf{95.55\eb{0.32}} & 93.11\eb{0.70} &  89.30\eb{0.21}   &  80.21\eb{1.91}  &  \textbf{93.78\eb{0.91}} \\	
	\midrule
\multirow{2}{*}{C2D (DivideMix with SimCLR)}  & \multirow{2}{*}{PreAct ResNet-18 }      & Peak  &96.43\eb{0.07}&95.32\eb{0.12}	&\textbf{94.40\eb{0.04}}&\textbf{93.57\eb{0.09}} &   \textbf{89.24\eb{0.75}}   & 93.45\eb{0.07}\\	            
			                &                               & Final  & 96.23\eb{0.09}  & 95.15\eb{0.16} &\textbf{94.30\eb{0.12}}  & \textbf{93.42\eb{0.09}}    &  \textbf{87.72\eb{2.21}}    &  90.75\eb{0.35} \\		
	\bottomrule
\end{tabular}
}
	\caption{
			Classification accuracy (\%, mean\eb{std} over five runs) on CIFAR-10. C2D achieves consistently high accuracy under different noise rates and types, with markedly improved performance under very-high noise conditions.  Meta-learning results provided by \citet{li2020dividemix}.
		}
	\label{tbl:cifar10}
\end{table*}

%% file: sections/040_experiments.tex
\input{sections/result_table_cifar_100.tex}

\section{Experimental results}
\label{sec:exp}

Our evaluation of the proposed framework uses two state-of-the-art LNL methods: ELR+ \cite{liu2020earlylearning} and DivideMix \cite{li2020dividemix}, both on synthetic and real noise. We follow common practice in synthetic noise benchmarks and use CIFAR-10 and CIFAR-100 \cite{krizhevsky2009cifar}, varying the amount of injected noise. For the real noise setting, we use WebVision \cite{li2017webvision}, a dataset of \textasciitilde2.4 million images based on queries generated from the 1,000 ImageNet \cite{ILSVRC15} classes, and Clothing1M \cite{xiao2015learning}, which contains \textasciitilde1 million images of 14 classes of clothing. Both datasets are acquired by web crawling.

Common evaluation using Clothing1M includes utilizing the ImageNet pre-trained network, while for CIFAR and WebVision, networks are trained from scratch. Thus, for the former we provide a comparison with supervised pre-training, while for the latter we compare self-supervised pre-training with no pre-training whatsoever.

\subsection{CIFAR-10 and CIFAR-100}
\label{subsec:cifar_exp}

We conducted experiments with two types of label noise: symmetric and asymmetric. 
Symmetric noise is generated by randomly replacing the
labels in a percentage of the training data with a random label drawn from a uniform distribution over all labels. 
Following the common approach \cite{zhang2016understanding,arazo2019mcorr,li2020dividemix}, the new label may be the real one. In this way, we are guaranteed that the clean label is the most frequent label for each class for any noise level. Thus, the real number of mislabeled examples is smaller by $\nicefrac{1}{n_{cl}}$. 
Asymmetric noise is designed to mimic the structure of real-world label errors, where classes that are generally similar in appearance are more likely to switch labels. In this case, we follow a scheme proposed by \citet{patrini2017making}. %: for CIFAR-10 there are particular pair of classes which may switch labels, while for CIFAR-100 noisy labels are chosen out of other labels from the same super-class, containing five classes.

\paragraph{Implementation details.}
For both methods, we followed the setup of the original experiments as close as possible. We also used the original architectures,   PreAct ResNet-18 \cite{he2015deep} for DivideMix and ResNet-34 for ELR+. Since self-supervised training is known to benefit from increased network capacity \cite{henaff2019cpc2, chen2020simclr}, for CIFAR-100 we performed experiments with ResNet-50 as well.
For self-supervised pre-training, we used a SimCLR implementation\footnote{\url{https://github.com/HobbitLong/SupContrast}} in PyTorch \cite{paszke2019pytorch}. The self-supervised model was trained for 1000 epochs on 4 NVIDIA 2080 Ti GPUs.

ELR+ required no hyperparameter tweaking. For DivideMix, we performed a number of minor modifications: (a) to accommodate ResNet-50 in GPU memory, we reduced the batch size from 128 to 64 and (b) for DivideMix we observed that our network kept improving after 300 epochs and thus increased training length to 360 epochs.  In addition, we tuned the hyperparameters mentioned in the original paper \cite{li2020dividemix}: the unlabeled loss weight $\lambda_{\mathcal{U}}$, the number of warm-up epochs, and the threshold for noisy label prediction $\tau$.
For $\lambda_{\mathcal{U}}$, we acquired similar results with and without C2D. Those results match the results of \citet{li2020dividemix}, except  that increasing $\lambda_{\mathcal{U}}$ also benefits the baseline DivideMix solution in high noise settings: for CIFAR-100 with 80\% noise, increasing $\lambda_{\mathcal{U}}$ from 150 to 500 improved DivideMix accuracy from 60.2\% to 61.3\%. 
As discussed in \cref{sec:warmup}, strong pre-trained features are expected to reduce the required warm-up duration. We found that five epochs were sufficient for CIFAR at all noise levels. As a reference, DivideMix uses 10 epochs for CIFAR-10 and 30 epochs for CIFAR-100. Lastly, we set the GMM threshold to $\tau=0.03$, which is significantly lower than the $0.5$ used by DivideMix. This can be explained by the fact our model is able to determine most of the noisy examples with high confidence. 

\paragraph{Results.}
\cref{tbl:cifar10} presents the comparison of our method with prior state of the art for symmetric and asymmetric noisy labels on the CIFAR-10 dataset. 
%Following \citet{li2020dividemix}, 
``final'' refers to the accuracy at the end of training for DivideMix, and the one with highest internal validation score for ELR+ as done in the original papers. ``peak'' refers to the highest validation score achieved during training. 

% We present accuracy at the end of training (denoted ``final'') together with the highest one achieved during training (denoted ``peak''). We follow the original approach of each of methods, taking the last epoch output for DivideMix and the one with highest internal validation score for ELR+ as the ``final'' result.

In addition to maintaining consistently high classification accuracy across all noise levels, C2D significantly outperforms prior methods at high noise levels  ($\geqslant$ 80\%). We attribute this desired behavior to the fact that our pre-trained features are agnostic to the noise level.  %When presented with asymmetric noise, both DivideMix and C2D have a degradation between peak and final accuracy. Even though C2D shows stronger degradation, it performed on par with previous art in terms of peak accuracy. 

\cref{tbl:cifar100} shows the classification accuracy on CIFAR-100. Compared with CIFAR-10, this task is more complex, resulting in a steeper drop in performance of prior methods as noise rates increase. In contrast, C2D demonstrates a graceful degradation, achieving a remarkable gain of more than 30\% in accuracy at 90\% noise level. We therefore decided to stress test C2D by subjecting it to an extreme noise level of 95\%. Despite a higher variance in the results (measured across five noise realizations), C2D still achieved a final accuracy of above 38\% (and at least 30\% in each individual run), surpassing the performance achieved by previous approaches at a noise rate of 90\%. 
In asymmetric noise, C2D performed similarly to prior art with ResNet-18, and achieved a minor improvement over ELR+ \cite{liu2020earlylearning} with larger networks (ResNet-34 and ResNet-50).

We also provide an additional baseline which uses only first stage of C2D, i.e., self-supervised pre-training followed by vanilla cross-entropy training with mixup. For harder tasks ($\geqslant$90\% noise on CIFAR-100) the improvement provided by second stage (ELR+ training) is marginal, while for intermediate noise rates the gain is maximal (e.g., 8\% difference for CIFAR-100 with 50\% noise and CIFAR-10 with 95\% noise).

\subsection{Clothing1M}

\begin{table}
	\centering
    \setlength\extrarowheight{2pt}
	\begin{tabular}	{l  c }
		\toprule	 	
			\textbf{Method } & \textbf{Test accuracy} \\
			\midrule			
			Cross-entropy & 69.21 \\
			F-correction \cite{patrini2017making}  &69.84\\	
			%M-correction \cite{arazo2019mcorr}&  71.00 \\			
			Joint-Optim \cite{tanaka2018joint}  & 72.16\\			
			MetaCleaner \cite{zhang2019metacleaner} & 72.50\\
			Meta-Learning \cite{li2019learning}  & 73.47\\	
			P-correction \cite{yi2019pencil}&73.49\\
			Self-learning \cite{han2019deepself} & 74.45\\
			DivideMix \cite{li2020dividemix}& 74.76\\
			ELR+ \cite{liu2020earlylearning}& \textbf{74.81}\\
			\midrule
			Cross-entropy (SimCLR init.)	 & 72.05\\
			Cross-entropy (BT init.)	 & 73.03\\
			C2D (ELR+ with BT) & 73.52\\
			C2D (DivideMix with SimCLR) & 74.30\\
			C2D (ELR+ with SimCLR) & \textbf{74.58\eb{0.15}}\\
		\bottomrule
	\end{tabular}
	%}
	\caption
		{
		%\small	
		Comparison with state-of-the-art methods in test accuracy (\%) on Clothing1M. The upper part of the table uses ImageNet pre-training, while the lower half does not. %Results for baselines are copied from original papers.
		}
	\label{tbl:clothing}
\end{table}

We tested our framework on the real-life noise present in the Clothing1M dataset \cite{xiao2015learning}.
As some of the  manually labeled images have both clean and noisy labels, we can estimate the noise level as approximately 38.5\%. We also use these double-labeled samples to compute noise-related metrics such as the ROC AUC of noise detection. 
Implementation details are specified in the appendix. 

\paragraph{Results.} 

The default approach \cite{patrini2017making,tanaka2018joint,zhang2019metacleaner,li2019learning,yi2019pencil,han2019deepself,li2020dividemix,liu2020earlylearning} for Clothing1M is to leverage a ResNet-50 pre-trained on ImageNet. The rich variety of visual concepts along with high-quality labels provides a strong initialization for network weights. C2D, on the other hand, uses only the dataset itself for pre-training. 
A comparison with state-of-the-art methods is reported in \cref{tbl:clothing}. The results highlight two interesting phenomena. First, by comparing the performance of the standard cross-entropy training, we confirm that our self-supervised pre-training is significantly better than ImageNet pre-training, demonstrating the warm-up gain C2D brings. 
A second observation is that, similarly to harder instances of CIFAR, this advantage is not leveraged by the LNL methods, resulting in an overall performance similar to ImageNet pre-training. This may be attributed to the complicated noise structure and leaves room for research of the way the methods utilize the improved initialization. Finally, it is encouraging that nearly state-of-the-art results (74.58\% vs.\ 74.81\% accuracy) can be achieved without external data.

% C2D achieves only a 0.5\% accuracy gap from the current state of the art. \ez{TODO}
% %Importantly, all the compared methods use ImageNet pre-trained features.
% As an additional baseline, we also provide results for cross-entropy fine-tuning of the same self-supervised network.

% \ez{REWRITE}
% \orl{let's discuss this after the final result is in.}
% As discussed in \cref{sec:warmup}, supervised pre-training may compensate for the (already minor) domain gap by eliminating the task gap, which may explain why C2D makes no additional gain. Moreover, C2D excels at high noise rates, which is not the case for Clothing1M. 
%Interestingly, though, C2D did show a 3\% improvement in the ROC-AUC score compared to baseline (81\% vs.\ 78\%). 
% This suggests that the self-supervised features helped in separation but are less suited for classification than the supervised ones explicitly trained for this purpose on a richer dataset. %We also emphasize that C2D is unique in that it did not require any additional external data. 
%\orl{I wonder what would happen if we trained simclr on clothing and imagenet together...}. 

\subsection{WebVision}
\input{sections/result_table_webvision.tex}

Following previous work \cite{chen2019understanding,jiang2019synthetic,li2020dividemix,song2020robust,liu2020earlylearning}, we evaluate our framework on the mini-WebVision 1.0 dataset \cite{li2017webvision}, which contains the first 50 classes of the Google image subset for a total of \textasciitilde61,000 images.
Implementation details are specified in the appendix.
\paragraph{Results}
As shown in \cref{tab:webvision}, C2D outperforms previous works on both the WebVision and ImageNet validation  sets by at least 3\% top-1 accuracy. Since we used a different network architecture, we also evaluated vanilla DivideMix with ResNet-50, reaching \textasciitilde1\% degradation of top-1 accuracy when compared to Inception-ResNet-v2.

%% file: sections/result_table_cifar_100.tex
\begin{table*}
	\centering
    \setlength\extrarowheight{2pt}
    \resizebox{\linewidth}{!}{
\begin{tabular}{@{\extracolsep{4pt}}l l l ccccc c@{}} 
\toprule	 	
	\multirow{2}{*}{\textbf{Method}}&\multirow{2}{*}{\textbf{Architecture}}   &&\multicolumn{6}{c}{\textbf{Noise rate}}\\ \cline{4-9}
  &       &       & \textbf{20\%} & \textbf{50\%} & \textbf{80\%} & \textbf{90\%} & \textbf{95\%} & \textbf{Asym. 40\%}  \\%  & \textbf{Asym. 60\%}
\midrule
Meta-learning   &  \multirow{2}{*}{PreAct ResNet-32 }    & Peak  & 68.5          & 59.2          & 42.4          & 19.5          &       --        &        --                 \\
\cite{li2019learning}&                      & Final  & 67.7          & 58.0          & 40.1          &14.3           &       --        &             --            \\
\midrule				
ELR+            &  \multirow{2}{*}{ResNet-34 }     & Peak  &      --        &              -- &        --       &	        --    &         --      &           --             \\
\cite{liu2020earlylearning}&                & Final  &  77.6          & 73.6          & 60.8          &33.4           &       --        &       77.5         \\
\midrule		
ODD	    &    \multirow{2}{*}{WRN-28-10 }         & Peak  &   79.1\eb{0.1}       &   --       &       --    &       --   &     --          &     --    \\	            
\cite{song2020robust}  &                    & Final  &     --     &    --       &       --    &  --       &        --       &          --             \\	          	
\midrule		
DivideMix	    &    \multirow{2}{*}{PreAct ResNet-18 }         & Peak  &  77.3         & 74.6          &61.6$^*$            &31.5           &      --         &    72.2$^*$      \\	            
\cite{li2020dividemix} &                    & Final  &   76.9        & 74.2          & 61.3$^*$              & 31.0          &       --        &           72.4$^*$             \\	   	
	\midrule		
\multirow{2}{*}{CE+mixup with SimCLR}  & \multirow{2}{*}{ResNet-34 }      & Peak  & 76.46\eb{0.15} & 69.14\eb{0.31}	 & 61.39\eb{0.26} & 55.51\eb{0.24} &  43.59\eb{0.59}  & 65.19\eb{0.63}\\	            
			                &                               & Final  & 76.34\eb{0.19} & 67.97\eb{1.22} & 60.81\eb{0.67}  &   54.64\eb{0.72}  &   42.11\eb{1.99} &  54.75\eb{0.93} \\	
	\midrule	
\multirow{2}{*}{C2D (ELR+ with SimCLR)}  & \multirow{2}{*}{ResNet-34 }      & Peak  & 79.18\eb{0.19}&	76.33\eb{0.31}& 64.72\eb{0.18}&55.08\eb{0.32} &  44.06\eb{0.84}  & \textbf{77.87\eb{0.29}}\\	            
			                &                               & Final  & 79.03\eb{0.20}  & 76.10\eb{0.36} & 64.18\eb{0.13} &   54.06\eb{1.50}   &  42.60\eb{1.67}  &  \textbf{77.63\eb{0.27}} \\	        	
\midrule		
\multirow{2}{*}{C2D (DivideMix with SimCLR) } &    \multirow{2}{*}{PreAct ResNet-18} 	& Peak & 78.69\eb{0.17}     & 76.43\eb{0.25}    &67.78\eb{0,30}      &58.70\eb{0,31}& 38.89\eb{1.19}&       75.48\eb{0.16}   \\	            
         &                    & Final                               & 78.32\eb{0.35}      & 76.07\eb{0.41}     &67.43\eb{0,30}      &58.45\eb{0,30}& 38.03\eb{2.13}&       75.06\eb{0.16}     \\	
\midrule		
\multirow{2}{*}{C2D (DivideMix with SimCLR) } &     \multirow{2}{*}{ResNet-50 }	& Peak & \textbf{81.60}      & \textbf{79.54}     &\textbf{71.65}      &\textbf{64.30}& \textbf{49.11} & \textbf{77.92}     \\	            
&&                                                                Final & \textbf{80.89}      & \textbf{79.20}     &\textbf{71.53}      &\textbf{63.91}& \textbf{48.50} & \textbf{77.78}     \\	
\bottomrule
\end{tabular}
}
	\caption{
			Peak and final classification accuracies (\%,  mean\eb{std} over five runs) on CIFAR-100. Unlike previous methods that suffer from rapid degradation, C2D was able to maintain good performance even under severe noise. %, achieving a remarkable $>30\%$ performance gain at $90\%$ noise rate.
			Meta-learning results provided by \citet{li2020dividemix}. $^*$ denotes results acquired by us based on published code.
		}
	\label{tbl:cifar100}
\end{table*}	

%% file: sections/result_table_webvision.tex
\begin{table*}
\centering
\setlength\extrarowheight{2pt}
% \resizebox{\linewidth}{!}{
\begin{tabular}{@{\extracolsep{4pt}}cllcccc@{}}
\toprule
\multirow{2}{*} & \multirow{2}{*}{\textbf{Method}} &\multirow{2}{*}{\textbf{Architecture}} & \multicolumn{2}{c}{\textbf{ILSVRC12}} & \multicolumn{2}{c}{\textbf{WebVision}} \\
                               &                                  &                                       & \textbf{Top-1}      & \textbf{Top-5}  & \textbf{Top-1}   & \textbf{Top-5}      \\
\midrule
% \multirow{4}{*}{Full}          &Vanilla                                    & Inception-ResNet-v2 & 61.7&82.4 & 70.9&88.0 \\
%                               & DCL \cite{saxena2019data}                 & ResNet-18 & --&-- &67.5&-- \\
%                               & MentorMix \cite{jiang2019synthetic}       & Inception-ResNet-v2 &  67.5&87.2 &  74.3&90.5 \\
%                               & C2D (ours)                                 &&  & & & \\
% \hline
\multirow{7}{*}          & MentorNet \cite{jiang2018mentornet}       & Inception-ResNet-v2 & 63.8&85.8 & -- & -- \\
                               & Iterative-CV \cite{chen2019understanding} & Inception-ResNet-v2 & 61.6&85.0 & 65.2&85.3 \\
                               & MentorMix \cite{jiang2019synthetic}       & Inception-ResNet-v2 &  72.9&91.1 & 76.0&90.2 \\
                               & DivideMix \cite{li2020dividemix}          & Inception-ResNet-v2 &  75.20&90.84 & 77.32&91.64 \\
                               & ODD \cite{song2020robust}                 & Inception-ResNet-v2 &  66.7&86.3 & 74.6&90.6 \\
                               & ELR+ \cite{liu2020earlylearning}          & Inception-ResNet-v2 &  70.29&89.76 & 77.78&91.68 \\
                               & DivideMix$^*$ \cite{li2020dividemix}      & ResNet-50 & 74.42\eb{0.29} & 91.21\eb{0.12} & 76.32\eb{0.36} & 90.65\eb{0.16} \\
                               & C2D (DivideMix with SimCLR)                                & ResNet-50 & \textbf{78.57\eb{0.37}} &  \textbf{93.04\eb{0.10}} & \textbf{79.42\eb{0.34}} & \textbf{92.32\eb{0.33}}\\
\bottomrule
\end{tabular}
%}
\caption{Accuracy (\%, mean\eb{std} over five runs) on the WebVision validation set and the ILSVRC12 (ImageNet) validation sets, for the networks trained on (mini) WebVision dataset. %Results for baselines are copied from original papers. 
$^*$ denotes results acquired by us based on published code.}
\label{tab:webvision}
\end{table*}

%% file: sections/050_details.tex
\section {Analysis}
\label{sec:analysis}

% In this section, we analyze the behavior of C2D in 4 aspects: \cref{subsec:warmup} focuses on warm-up performance for different pre-training strategies, \cref{subsec:super_comparison} evaluates the supervised pre-training as an alternative to C2D. 

% \cref{subsec:ssl_gap} measures the performance gap between C2D and (noise free) semi-supervised learning, and 

\subsection{Warm-up performance}
\label{subsec:warmup}
In \cref{sec:warmup}, we defined low feature quality and poor loss separation as a major obstacle to improving LNL performance. Here, we show that C2D improves both metrics.
%To assess the performance of the warm-up phase with and without C2D, we analyze the feature quality and the loss separation of 
%Self-supervised pre-training serves a dual purpose: providing a better initialization for the classification task while preventing memorization of noisy labeled samples. In the following, we analyze these properties. First,
We start with a qualitative comparison of the features learned on the CIFAR-10 data with and without C2D at the end of the warm-up. We do this by visualizing the features in \cref{fig:umap} using a dimensionality-reduction technique \cite{lel2018umap}. % colored using the ground-truth labels
C2D features (upper row) are clearly better clustered and easier to separate than the baseline (bottom row) at both 20\% and 90\% noise levels. 
Furthermore, at high noise rates, the baseline features suffer from acute degradation (note the blending of the dog and deer categories into their surrounding classes), while C2D features maintain some fidelity.

\begin{figure*}
\centering
        \includegraphics[width=\linewidth]{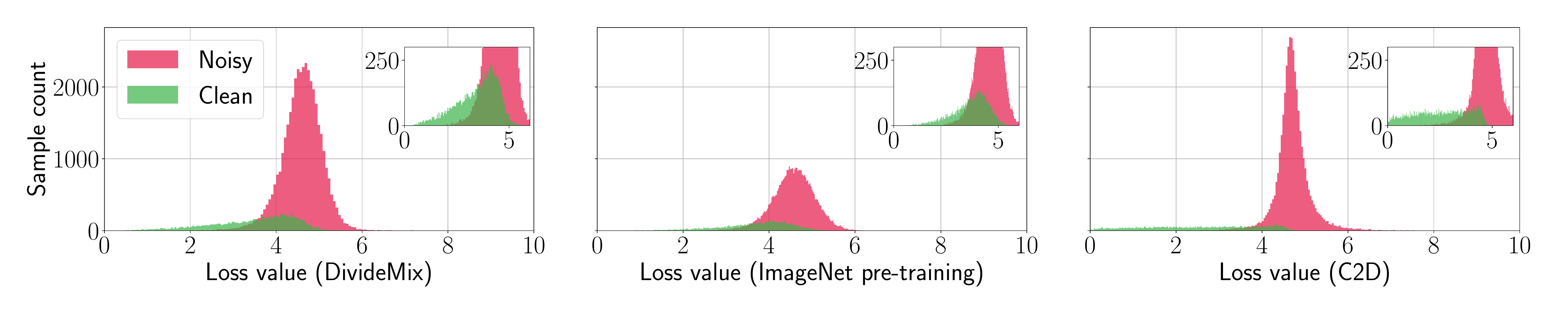}
 \caption{Loss distribution of clean and noisy samples after warm-up on CIFAR-100 with 80\% noise for DivideMix, DivideMix with ImageNet pre-training, and C2D. As seen in the zoom-in, ImageNet pre-training damages the separability whereas self-supervised pre-training (C2D), improves it. }
\label{fig:loss_distribution}
\end{figure*}

To evaluate the impact of different pre-training approaches (a standard warm-up, ImageNet pre-training, and C2D) on the performance of the warm-up stage,  in \cref{fig:tradeoff} we visualize the properties defined in \cref{sec:warmup}: ROC-AUC score of noise detection and the linear accuracy using clean labels under different noise levels for CIFAR-100. 
%To evaluate the quality of noise detection, in TODO we present the ROC-AUC score of noise detection and the effective noise rate, defined as the share of noisy samples in the labeled part of the dataset. 
C2D demonstrates significantly higher overall feature quality and loss separability at all noise levels: even after one epoch both values are higher than the peak ones for vanilla warm-up and ImageNet pre-training. Moreover, thanks to the pre-training on the same dataset and discarding the labels, C2D shows faster convergence and smaller performance degradation with increased noise.
In addition,  in \cref{fig:loss_distribution} we visualize a histogram of loss values of clean and noisy samples for all three approaches. % supports those claims. 
It can be seen that  C2D does indeed provide a lower loss for most clean samples with the smallest overlap between clean and noisy samples, which allows both faster convergence and higher overall performance.

\subsection{The drawbacks of supervised pre-training}
\label{subsec:super_comparison}

% imagenet pre-training helped once and hurt once. C2D was consistent.
% needs external data. 

Utilizing supervised pre-trained features from large cleanly labeled data source (e.g., ImageNet \cite{alex2019bit}) is standard procedure in noise-free learning tasks. In the context of LNL, much less is known about ImageNet pre-training benefits besides its high performance on Clothing1M. Therefore,  we chose to apply ImageNet pre-training to noisy CIFAR.

We ran ELR+ and DivideMix on CIFAR-100 with a network initialized with ImageNet pre-trained weights. 
One may expect that a small domain gap along with versatile high-quality pre-trained features will make this an almost ideal setup.  Indeed,  \cref{fig:tradeoff} shows, at least for 90\% noise level, notable improvement in the warm-up phase, which is, however, far inferior to that of C2D.

Surprisingly, these improvements did not result in an improvement of the performance of LNL.
On ELR+ \cite{liu2020earlylearning},  adding ImageNet pre-training reduced the accuracy from 60.8\% to 48.58\eb{0.16}\% for CIFAR-100 with 80\% noise and from 33.4\% to 23\% with 90\% noise.
On DivideMix, in addition to an expected shortening of the required warm-up length (from 30 to 10 epochs), at the end of the warm-up, on 80\% noise we observed an increase both in the ROC-AUC score 
and the classification accuracy. Yet, most concerning was the almost immediate failure of DivideMix when entering the second stage of training.
After the warm-up, the loss values of the clean and noisy samples were almost indistinguishable, which resulted in a severe decrease in classification accuracy as depicted in \cref{{fig:loss_distribution}}. Despite our attempts to rectify this behavior, this phenomenon persisted across various sets of hyperparameters.
%threshold values ranging from 0.03 (C2D) to 0.5 (DivideMix), using either fixed or linearly increased values.  

% our experiments reveal that ImageNet pre-trained features do not necessarily aid learning. In particular, in CIFAR, aside from damaging the warm-up stage, the final performance degraded compared to training from scratch. On the other hand, on Clothing1M this pre-training was instrumental to getting state-of-the-art results. 

% While analysing the conditions that make supervised pre-training succeed (e.g. domain gap, noise level, etc.) is beyond the scope of this paper, our results indicate that its effect on LNL is generally unpredictable. On the contrary, using C2D resulted in consistent improvement across all our experimental settings. In addition, C2D does not require external data nor additional supervision.
Our results indicate that supervised pre-training has a generally unpredictable effect on LNL. We leave the influence analysis of different conditions (e.g. domain gap or noise level) to future work. On the contrary, using C2D resulted in consistent improvement across all our experimental settings. In addition, C2D does not require external data nor additional supervision.

%, we believe this method should be widely adopted. 
%we suspect that the fast-adaptation property of the pre-trained network may have damaged the network's resilience to noise. 

%While a full analysis of supervised transfer learning under noise conditions is outside the scope of this work, 

\subsection{Self-supervised pre-training method}
\label{subsec:selfsup_method}
Further verifying the universality of our approach, we examine how the self-supervised pre-training approach affects the performance of C2D. In addition to SimCLR, we used Barlow Twins \cite{zbontar2021barlow}, due to its high performance and the distinct differences with SimCLR. The results are presented in \cref{tbl:clothing}. Using cross-entropy fine-tuning (without accounting for noise) Barlow Twins outperforms SimCLR, achieving only 1.5\% below the state-of-the-art. While this advantage did not translate into significantly better overall performance (0.5\% improvement when using ELR+), high overall performance without supervised pretraining indicates that C2D can work well with other self-supervised feature extraction techniques. 

\subsection{Gap between LNL and semi-supervised learning} 
\label{subsec:ssl_gap}

\begin{table}
	\centering
    \setlength\extrarowheight{2pt}
	\begin{tabular}{@{\extracolsep{4pt}}l cc@{}} 
		\toprule	 	
		  \multirow{3}{*}{\textbf{Method}}& \multicolumn{2}{c}{\textbf{Missing/noisy}}\\ &\multicolumn{2}{c}{\textbf{label rate}} \\ \cline{2-3}
		  & \textbf{80\%}  & \textbf{90\%}\\	\midrule
			MixMatch	 & 70.46 & 64.60\\	
			MixMatch (SimCLR init.)	  &71.86 & 66.10\\
			%DivideMix	& Best & &  & &\\
			C2D (DivideMix with SimCLR) & 71.65 & 64.30 		\\
		\bottomrule
\end{tabular}
	\caption
		{
			C2D nearly closes the gap with  semi-supervised training on the same clean set size.  
		}
	\label{tbl:cifar-semi}
\end{table}

In the case of DivideMix \cite{li2020dividemix} and other methods that utilize semi-supervised learning, semi-supervised accuracy is effectively the upper bound on the performance.
A significantly better noise separation ability along with the improved initialization raises the question whether any performance gap remains between LNL and semi-supervised learning.  
To answer this question, we compared the performance of C2D (with DivideMix)
with MixMatch -- a semi-supervised method -- provided with the same amount of labels as the clean portion of the C2D training set. This procedure is roughly equivalent to replacing the noise detection procedure with an oracle. The result for 80\% and 90\% noise levels in CIFAR-100 are reported in  \cref{tbl:cifar-semi}.  Remarkably, C2D is on par with MixMatch and less than 2\% below MixMatch with self-supervised pre-training. Even though the LNL setup has \textit{strictly less information} than the semi-supervised one, these results indicate that good features can compensate for this lack of information even under severe noise conditions.

%% file: sections/060_conclusion.tex
\section{Conclusion}
\label{sec:conclusion}

In this paper, we proposed Contrast to Divide (C2D), a simple yet powerful framework for learning with noisy labels that do not rely on external labeled data and leverages self-supervised pre-training instead. 
We have identified and analyzed a major obstacle to LNL: due to memorization, loss separability and feature quality %noisy labels uring 
after warm-up are bounded, and deteriorate quickly with increasing noise level. 
%to bad loss separability and low quality of extracted features. 
Moreover, while the natural robustness of the neural networks allows us to acquire good results, little is known about the sources of this robustness and how to improve it. We have shown that self-supervised pre-training boosts both warm-up goals, which in turn dramatically improves the performance of existing LNL approaches. 

% use it. it's easy. more trustworthy then supervised pre-training. 
C2D is straightforward to implement,  does not require any external data, and works out of the box with multiple existing LNL approaches, demonstrating consistently high performance across various noise levels. In real-life settings, we tested C2D on mini-WebVision and achieved more than a 3\% top-1 accuracy boost over the previous state of the art. In addition, C2D shows stable performance under severe noise,  outperforming prior art by more than 20\% for 90\% noise on CIFAR-100 and nearly closing the gap with semi-supervised learning trained on the same amount of labeled samples as the clean portion. 

Even though C2D provides significant performance improvement, studying the robustness property is still an open research question.  Clothing1M results also suggest that the way existing methods utilize pre-trained features can be improved too. We leave those questions for the future work.

%% file: sections/070_appendix.tex
\appendix{}

%%% FIGURE NUMBERING IN APPENDIX
\renewcommand\thefigure{\thesection.\arabic{figure}} 
\renewcommand\thetable{\thesection.\arabic{table}} 
\renewcommand\theequation{\thesection.\arabic{equation}}  
\setcounter{figure}{0}  
\setcounter{table}{0}

\crefalias{section}{appsec}
\crefalias{subsection}{appsec}
\crefalias{subsubsection}{appsec}

\section{Implementation details}
\subsection{Clothing1M}

As most previous works, we used ResNet-50 architecture, but did not utilize ImageNet pre-training.
For self-supervised pre-training, we used a SimCLR implementation\footnote{\url{https://github.com/HobbitLong/SupContrast}} in PyTorch \cite{paszke2019pytorch}, trained on 8 NVIDIA 2080 Ti GPUs for 750 epochs. We trained the network using the AdamW optimizer \cite{loshchilov2017decoupled}.

\paragraph{DivideMix}
For DivideMix, we used a weight decay of $0.001$, and a batch size of $32$. As in the case of CIFAR, the warm-up period is five epochs. We trained the network for 120 epochs, with initial learning rate of $0.002$, reduced by a factor of 10 after 40 epochs. For each epoch, we sampled 1000 mini-batches from the training data with same amount of samples of every class (according to noisy label). 
We set $\lambda_{\mathcal{U}}=0$. Since a large amount of data is available, we found that increasing value of the threshold to $\tau=0.7$ improves the performance of the network.

\paragraph{ELR+}
For ELR+, we used the default hyperparameters, except for reduced learning rate ($0.001$).
\subsection{WebVision}

\paragraph{DivideMix}
For WebVision, we also used ResNet-50 architecture.
For self-supervised pre-training, we used a SimCLR implementation\footnote{\url{https://github.com/HobbitLong/SupContrast}} in PyTorch \cite{paszke2019pytorch}, trained on 8 NVIDIA 2080 Ti GPUs for 1000 epochs.
We trained the network using the AdamW optimizer \cite{loshchilov2017decoupled} with a weight decay of $0.001$, and a batch size of $32$. The warm-up period is one epoch. We trained the network for 80 epochs, with initial learning rate of $0.002$, reduced by a factor of 10 after 40 epochs.
We set $\lambda_{\mathcal{U}}=0$.

\section{Noise detection analysis}

\begin{figure*}
\centering
    \begin{subfigure}{0.49\linewidth}
        \includegraphics[width=\linewidth]{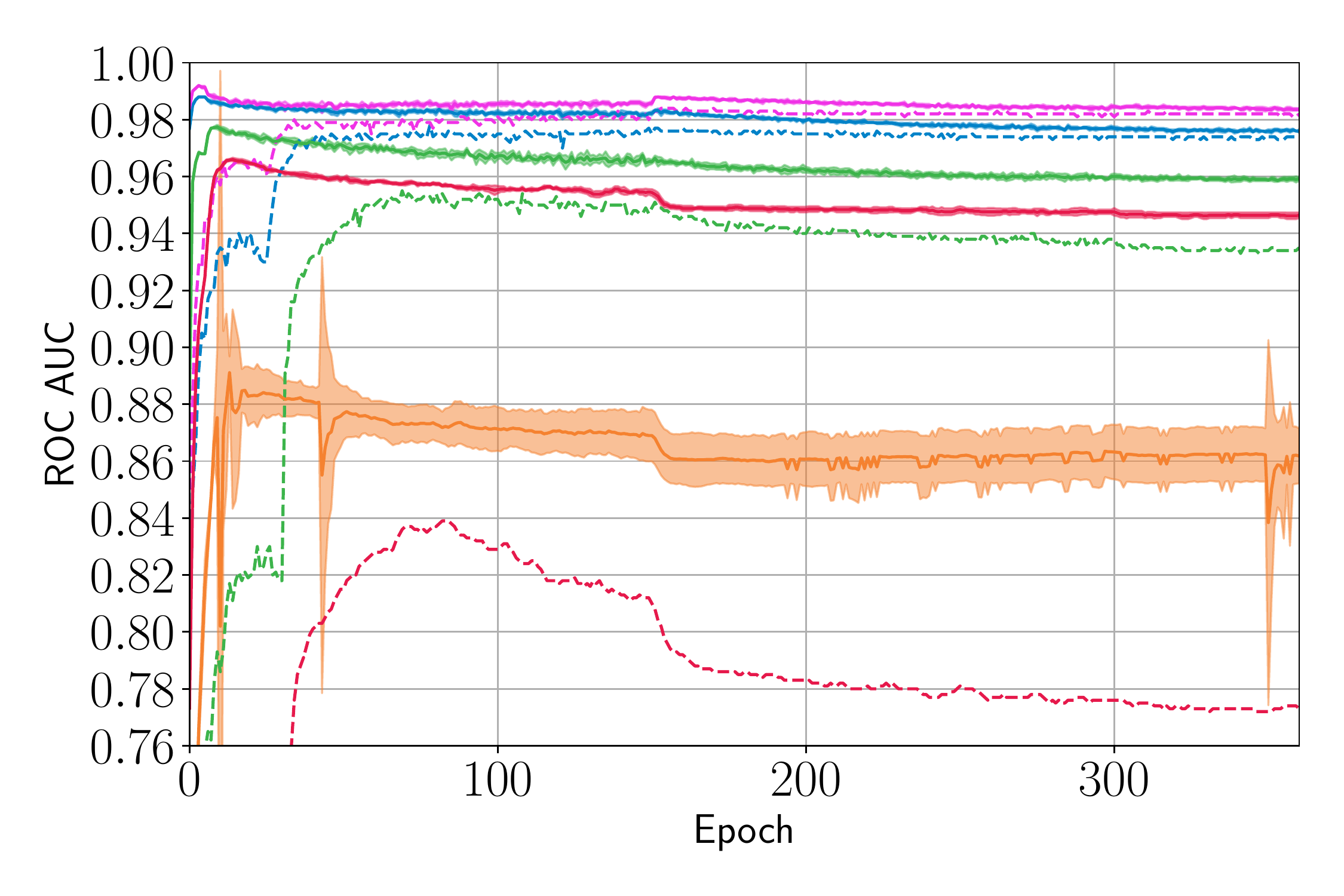}
    \end{subfigure}
    \begin{subfigure}{0.49\linewidth}
        \includegraphics[width=\linewidth]{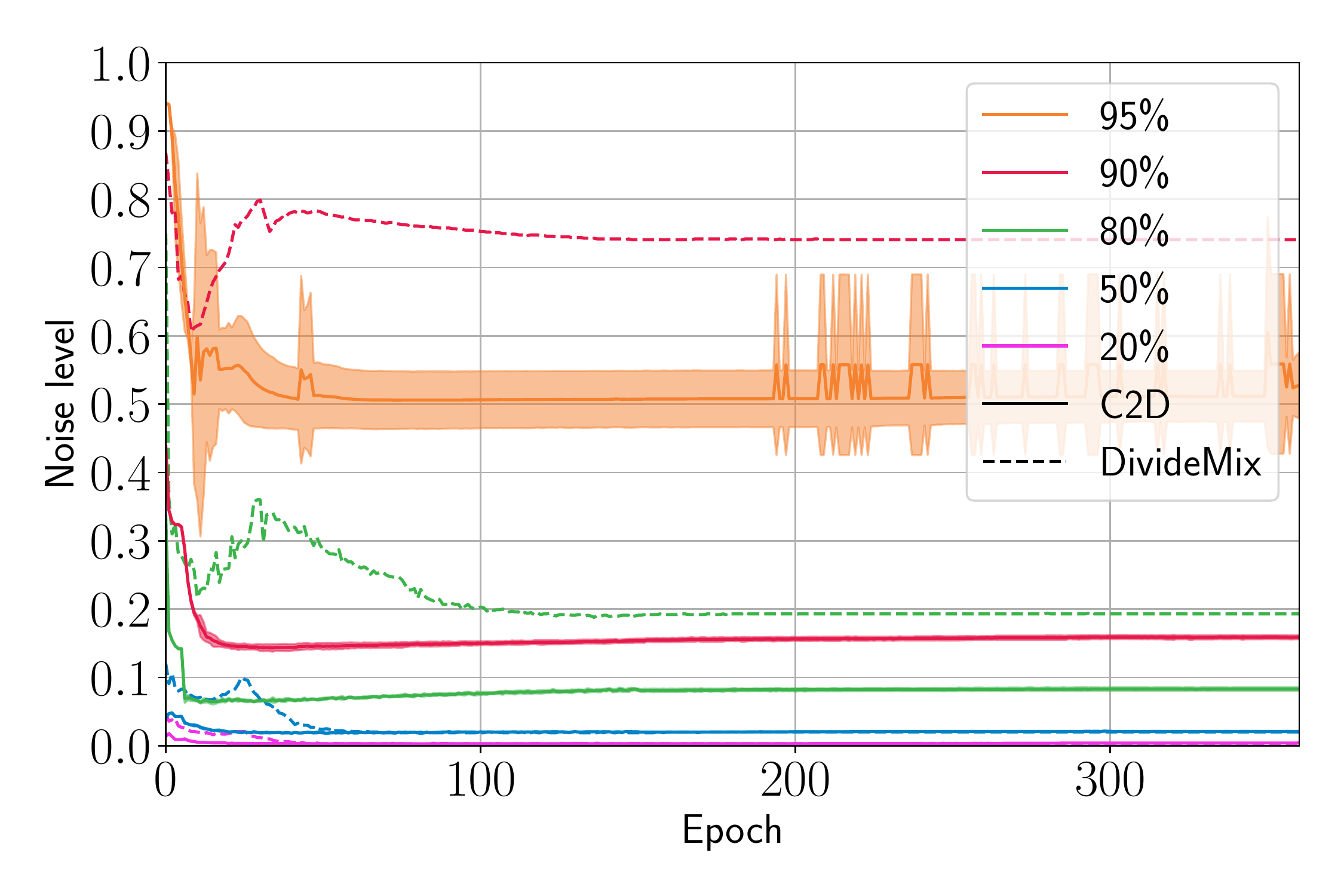}
    \end{subfigure}
 \caption{Training time ROC-AUC scores (left) and effective noise rates (right). C2D demonstrates higher initial score, faster rise, and more stable decrease in effective noise level.}
\label{fig:auc}
\end{figure*}

To evaluate the quality of noise detection, in \cref{fig:auc} we present the ROC-AUC score of noise detection and the effective noise rate, defined as the share of noisy samples in the labeled part of the dataset. C2D demonstrates multiple desired properties including a higher initial score, a much faster rise in separability score as well as a more stable decrease in effective noise level, and eventually a higher overall score and lower noise level. Moreover, even though C2D and the baseline both suffer from %inevitable 
decrease in the ROC-AUC score due to overfitting, C2D demonstrated a lower gap between the peak and final scores than the baseline.